\documentclass[letterpaper, journal, 10pt]{IEEEtran}
\usepackage{subfigure}
\usepackage{comment}
\usepackage{booktabs}
\usepackage{multirow}
\usepackage{latexsym}
\usepackage{algorithm}
\usepackage{algpseudocode}
\usepackage{subfigure}
\usepackage{stfloats}
\usepackage{graphicx}
\usepackage{listings}
\usepackage{xcolor}
\usepackage{amsmath}

\definecolor{codegreen}{rgb}{0,0.6,0}
\definecolor{codegray}{rgb}{0.5,0.5,0.5}
\definecolor{codepurple}{rgb}{0.58,0,0.82}
\definecolor{backcolour}{rgb}{0.95,0.95,0.92}
\lstdefinestyle{mystyle}{
    backgroundcolor=\color{backcolour},   
    commentstyle=\color{codegreen},
    keywordstyle=\color{magenta},
    numberstyle=\tiny\color{codegray},
    stringstyle=\color{codepurple},
    basicstyle=\ttfamily\footnotesize,
    breakatwhitespace=false,         
    breaklines=true,                 
    captionpos=b,                    
    keepspaces=true,                 
    numbers=left,                    
    numbersep=5pt,                  
    showspaces=false,                
    showstringspaces=false,
    showtabs=false,                  
    tabsize=2
}
\begin{document}
\title{Hyperdimensional Computing vs. Neural Networks: Comparing Architecture and Learning Process}

\author{
	Dongning Ma, \textit{Student Member}, IEEE, Xun Jiao, \textit{Member}, IEEE
	\thanks{Authors are with the Department of Electrical and Computer Engineering of Villanova University, Villanova, PA 19085. (e-mail: \{dma2, xun.jiao\}@villanova.edu)}
}

% \author{Dongning Ma, Xun Jiao*} 
% \affiliation{ 
%   \institution{Department of Electrical and Computer Engineering, Villanova University}
%   %\streetaddress{}
%   \city{Villanova} 
%   \state{PA 19085}
%   \country{USA}
% }
% \email{{dma2, xjiao}@villanova.edu }
\maketitle

\begin{abstract}
Hyperdimensional Computing (HDC) has obtained abundant attention as an emerging non von Neumann computing paradigm. Inspired by the way human brain functions, HDC leverages high dimensional patterns to perform learning tasks. Compared to neural networks, HDC has shown advantages such as energy efficiency and smaller model size, but sub-par learning capabilities in sophisticated applications. Recently, researchers have observed when combined with neural network components, HDC can achieve better performance than conventional HDC models. This motivates us to explore the deeper insights behind theoretical foundations of HDC, particularly the connection and differences with neural networks. In this paper, we make a comparative study between HDC and neural network to provide a different angle where HDC can be derived from an extremely compact neural network trained upfront. Experimental results show such neural network-derived HDC model can achieve up to 21\% and 5\% accuracy increase from conventional and learning-based HDC models respectively. This paper aims to provide more insights and shed lights on future directions for researches on this popular emerging learning scheme.
\end{abstract}

\begin{IEEEkeywords}
hyperdimensional computing, neural network, non von Neumann computing, bio-inspired computing
\end{IEEEkeywords}

\section{Introduction}
In recent years, machine learning has achieved tremendous success in a diverse range of domains, even surpassing the capability of human beings. However, such huge progress on machine learning comes with the drastically growing model complexity and the ever-increasing computation resources~\cite{sze2017efficient}. On the other hand, as machine learning has been continuously pushed to the edge such as embedded systems or (near-)sensor devices, the computation resources allocated for learning is significantly less than centralized servers or data centers~\cite{plastiras2018edge}. %To mitigate this issue, various methods or architectures that are specifically focused on enhancing the energy efficiency for edge intelligence have been proposed, including approximate computing techniques~\cite{mittal2016survey, han2013approximate}, specified architectures~\cite{chen2016diannao, shafique2017adaptive} and near-threshold computing~\cite{dreslinski2010near}.
%In addition / for existing machine learning algorithms
To address this challenge, researchers seek for non-conventional computing paradigms, and Hyperdimensional Computing (HDC), is one example seen as a promising alternative of conventional machine learning models~\cite{kleyko2021survey}. The key idea of HDC is to leverage the computing capabilities of distributed representation of high dimensional numerical vectors referred to as Hypervectors (HV)~\cite{kanerva2009hyperdimensional}. 
%The concept of HDC emerges from the hypothesized functioning of brain where, while both the brain and computer can perform similar recognition or learning tasks, there is a huge disparity between the architectures of each. 
Specifically, HDC is formulated to leverage HVs and their associated vector arithmetic to present, represent, and process information from different modalities~\cite{thomas2020theoretical}. Related studies show that HDC is able to achieve success on a diverse set of applications, including speech recognition~\cite{imani2017voicehd}, human activity recognition~\cite{ma2021molehd, kim2020geniehd}, NLP~\cite{thapa2021spamhd, liu2022l3e} and anomaly detection~\cite{wang2021brief}. In addition, various acceleration techniques are also proposed and implemented on heterogeneous platforms including GPGPU~\cite{kim2020geniehd}, FPGA~\cite{salamat2019f5} and even in-memory architectures~\cite{karunaratne2020memory}.

Although achieving energy efficiency and acceleration, one major obstacle preventing HDC to be applied for a broader range of applications is its relatively lower learning capability. For example, for a very basic benchmark of MNIST, the baselines of HDC solutions can only achieve less than 95\% accuracy~\cite{ma2021hdtest, duan2022lehdc}, which is much lower than a basic LeNet-like network~\cite{lecun1998gradient}. On the other hand, the application datasets which HDC shows advantage on are mostly small and relatively easier to differentiate. Therefore, such a limited set of applications when evaluating HDC algorithm is unlikely to justify the benefits of HDC and enable its practical use.

On the other hand, researchers endeavor to introduce techniques from neural networks into HDC for performance enhancement. For example, a fixed random connection neural network layer can be added into the HDC flow to achieve better performance~\cite{diao2021generalized}. Perceptrons specific for HDC are also proposed and further combined with other techniques such as drop-out~\cite{calimera2018exploring}. HDC is also found similar to binary neural networks based on which loss functions are also defined so that HDC can be trained with back propagation~\cite{duan2022lehdc}. Those related works reveals deeper connections between HDC and neural network and also motivate us to further explore the HDC foundations. This paper, specifically, present a comparative study to show the similarities and differences between HDC and neural networks. The main contributions are as follows:

\begin{itemize}
    \item We first present a recap of HDC preliminaries applied in the majority of related works, which we recognize as the ``canonical'' HDC flow. We then make a comparison between the HDC model and a two-layer ``neural network''\footnote{Typically, neural networks have at least 3 layers, however we recognize this two layers of perceptrons as a ``neural network'' for explanatory purposes and use this term throughout the paper.}, including both the architecture and the learning process.
    \item Experiments using two datasets of MNIST and CIFAR-10 show that HDC models can be directly transformed from a neural network which outperforms SOTA HDC models by 5\% to 21\% in accuracy. This paper provides an alternative angle to assess the learning capabilities of HDC and insights on future directions of enhancing and optimizing this promising computing scheme.
\end{itemize}

\section{Recap of HDC}
%In this section, we present a recap of the basic concepts in HDC such as the notions like HVs, HDC operations, HDC memories and similarity metrics, as well as the canonical flow of using HDC to develop a classifier including encoding, training, retraining and inference.
\subsection{Notions of HDC}
\label{sec:hdc_notion}

\subsubsection{Hypervector}
Hypervector (HV) is the fundamental ``building block'' of HDC. HVs have three most important properties. First, HVs are high dimensional numerical vectors, which are usually higher than 10,000 dimensions where each dimension is a number. This provides an extremely large space for HVs to represent information. 
%For example, if each number in a 10,000-dimensional HV is an 8-bit unsigned integer, then the number of possible HVs can achieve $2^{8^{10000}}$. 
Second, the HVs are usually randomly initialized and the numbers generated follows i.i.d. randomness. Because of the high dimensionality of HVs, this can ensure that two randomly initialized HVs can be (almost) orthogonal to each other. 
Third, HVs are holographic which means that within an HV, all the dimensions are recognized equally as to their contributions and there is no any dimension that is more important than others. In other words, the HV needs to be treated as a whole and not micro-coded. An HV of $D$ dimensions can be denoted as Eq.~\ref{eq:hv}, where $v_d$ is the number at the $d$-th dimension.

\begin{equation}
    \vec{V} = (v_1, v_2, ... , v_D)
    \label{eq:hv}
\end{equation}

\subsubsection{HDC operations}
% , 
There are three types of HV operations that are most frequently used in HDC: addition, multiplication and permutation and each type of operation has its specific physical meaning:
\begin{itemize}
    \item Addition $+$: takes two HVs as operands and performs element-wise addition on the numbers at the same dimension. Addition is usually used to aggregate the information of two HVs from the same modality and create a superposition of them.
    \item Multiplication $\times$: also takes two HVs as operands but performs element-wise multiplication on the numbers at the same dimension. Contrary to addition, multiplication is usually used to combine information from different modalities and create new information of another modality based on these two.
    \item Permutation $\rho$: only takes one HV as the operand and perform cyclic rotation (shift) over the HV. The shift amount can be configured based on specific applications. Permutation is usually used to reflect temporal or spatial patterns of information.
\end{itemize}

Note that all the three HV operations do not modify the dimension of HV operands, i.e., the input and output HV of each operation are in the same dimension.

% \begin{equation}
%     \begin{aligned}
%         & \vec{V_{a + b}} = \vec{V_a} + \vec{V_b} = (V_{a1} + V_{b1}, V_{a2} + V_{b2}, ... , V_{aD} + V_{bD}) \\
%         & \vec{V_{a \times b}} = \vec{V_a} \times \vec{V_b}  = (V_{a1} \times V_{b1}, V_{a2} \times V_{b2}, ... , V_{aD} \times V_{bD}) \\
%         & \vec{V_{\rho_1(a)}} = (V_{aD}, V_{a1}, ... , V_{aD - 1})
%     \end{aligned}
%     \label{eq:hdc_op}
% \end{equation}

\subsubsection{HDC memories}
Memories in HDC are a specific cluster of HVs which serve different functions in developing a model. Specifically, there are two types of memories: item memory and associative memory. Item memory is related to the input data, accommodates item HVs that are generated based on the input features. Assume the input sample has $M$ modalities of features and each feature can have $N$ possible values, then the item memory of each modality of feature can be generated as Eq.~\ref{eq:im}. On the other hand, associative memory is related to the output of the model, namely making predictions. If the classification task has $C$ classes, then the associative memory is configured as Eq.~\ref{eq:am}, in which each HV $\vec{A_c}$ inside the memory is the representation of a class. Associative memory is usually initialized with zero numbers.

\begin{equation}
\mathbf{I}_m = \{\vec{I_1}, \vec{I_2}, ... \vec{I_N} \}
\label{eq:im}
\end{equation}

\begin{equation}
\mathbf{A} = \{\vec{A_1}, \vec{A_2}, ... \vec{A_C} \}
\label{eq:am}
\end{equation}

\subsubsection{Similarity metrics}
Since each HV represents a specific information, there is a natural need of metrics that could represent the similarity between information that two HV respectively accommodate. Hamming distance and cosine similarity are the two mostly used similarity metrics. Hamming distance is for calculating the similarity between binary or bipolar HVs while cosine similarity can be used for HVs in different data types. We show the calculation of cosine similarity $\delta_{cos}$ in Eq.~\ref{eq:cosim} as an example. A higher similarity indicates that the two HVs compared share more information in common. 

\begin{equation}
\begin{aligned}
    \delta_{cos}(\vec{V_a}, \vec{V_b}) =  \frac{\vec{V_a} \cdot \vec{V_b}}{||\vec{V_a}||\times||\vec{V_b}||}
\end{aligned}
\label{eq:cosim}
\end{equation}

\subsection{HDC Model Development}
\label{sec:development}
A canonical flow of developing an HDC model features four major phases: Encoding, Training, Inference, and Retraining.

\subsubsection{Encoding}
Encoding is the basic phase of HDC model development. During encoding, the input samples are ``encoded'' into their representative HVs using a set of application-dependent HD operations $\Phi$ and the item memories. For example in Eq.~\ref{eq:encode}, assume the input sample has $M$ features: $\vec{F} = \{f_1, f_2, ... , f_M\}$, the values of each feature $f_m$ is used as indices to fetch corresponding item HV in the item memories $\mathbf{I}_m$. After encoding, the realistic features are now in the form of high-dimensional representations, namely the HVs which are used in all the other three phases of model development.

\begin{equation}
\vec{V} = \Phi(\mathbf{I}_1.index(f_1), \mathbf{I}_2.index(f_2), ... ,  \mathbf{I}_M.index(f_M))
\label{eq:encode}
\end{equation}

\subsubsection{Training}
Training is the phase where associative memory is trained using the encoding HVs from the training samples. As noted in Eq.~\ref{eq:training}, all the sample HVs sharing the same label $c$ are summed to the corresponding class HV $\vec{A_c}$ in the associative memory. This is to collect and aggregate information to build a representative HV for each class.

\begin{equation}
\mathbf{A} = \{ \sum{\vec{A_1}}, \sum{\vec{A_2}}, ... , \sum{\vec{A_C}}\}
\label{eq:training}
\end{equation}

\subsubsection{Inference}
Inference is the phase where associative memory is used to predict the class of unseen test samples. During inference, the unseen samples are still encoded using the same item memory as training. The encoded HV of the unseen sample is often referred to as the query HV $\vec{V_?}$. As shown in Eq.~\ref{eq:inference}, the similarity between query HV and every class HV in the associative memory is calculated and the class with the highest similarity is then determined as the predicted label $l$ for this unseen sample.

\begin{equation}
l = \delta_{cos}(\vec{V_?}, \mathbf{A}).argmax() 
\label{eq:inference}
\end{equation}

\subsubsection{Retraining}
Retraining is an optional phase of fine-tuning the trained associative memory to achieve higher performance of model. During retraining, the model iterates over the encoded training samples and make inference to obtain the prediction labels using the trained associative memory. Then the prediction label is compared with the ground truth label to identify if there is a discrepancy. If so, the associative memory is updated as Eq.~\ref{eq:retraining} describes: the encoded HV of the sample is subtracted from the class HV of the wrong prediction and added to the correct class. This is to reduce the wrong information from the mis-classifications and instead enhance the information of the correct class.

\begin{equation}
\begin{aligned}
 &      \vec{A}_{wrong} = \vec{A}_{wrong} - \vec{V} \\
  &  \vec{A}_{correct} = \vec{A}_{correct} + \vec{V}
\end{aligned}
\label{eq:retraining}
\end{equation}

In the following sections, we depart from the canonical notions of HDC as well as flow of developing an HDC classifier. Instead, we make comparisons on both the architecture and the learning process between a two-layer neural network and the HDC model and show that an HDC model can be potentially derived from a trained network as such.

\section{HDC versus NN: Architecture}
\subsection{Item Memory vs. Input Layer}
\begin{figure}
    % \centering
    \includegraphics[width = 1\columnwidth]{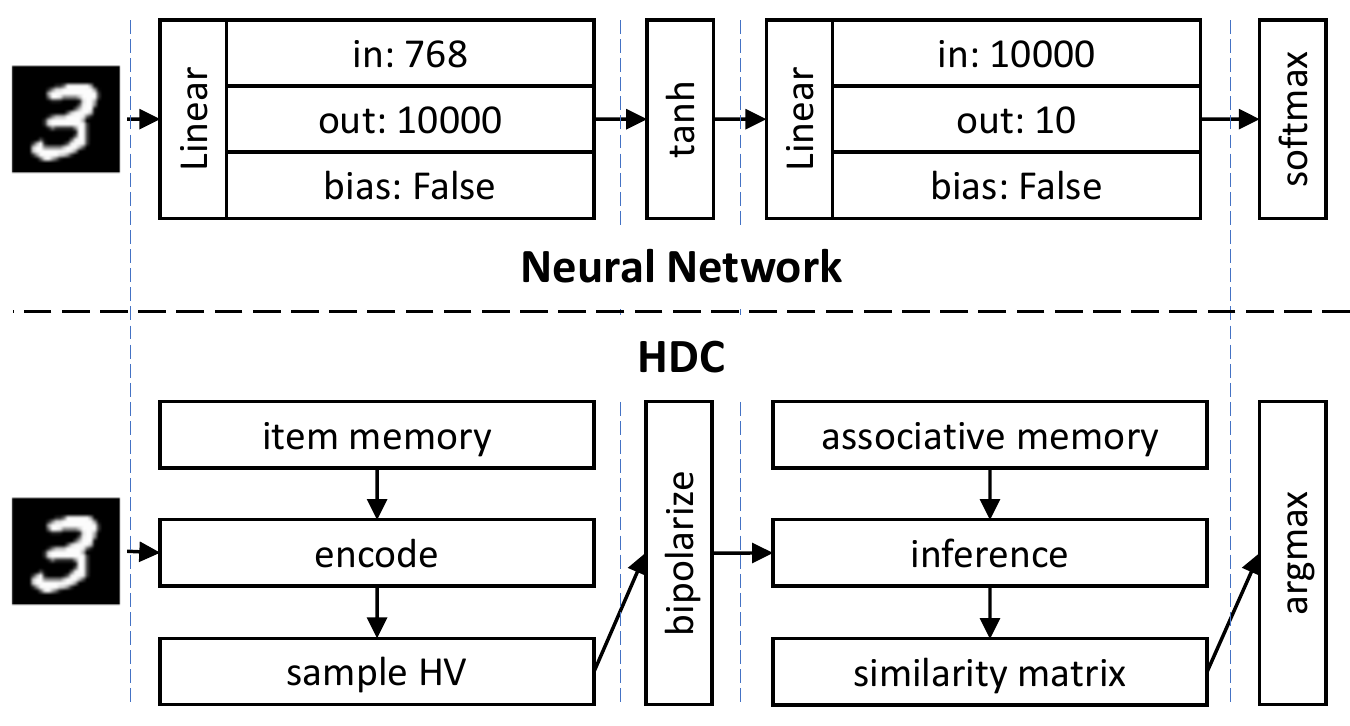}
	\caption{Architectural Comparison between HDC and neural network}
    \label{fig:network-arch}
\end{figure}
We first formulate the item memory and the encoding process as the processing of the input layer inside a neural network. As noted in Sec.~\ref{sec:development}, the input to the encoding phase is the sample and the output of is the HV with a specific (pre-defined) dimensionality. This resembles the classical input layer of a neural network that takes the samples then produce intermediate layer outputs. The encoding is conventionally application or task dependent, however the set of the HDC operations mostly features the three types of HDC operations as introduced in Sec.~\ref{sec:hdc_notion}. Therefore, such a set can be constructed by a combination of linear mappings, namely the forward pass of a linear layer.

Without loss of generality, we assume the encoding process uses the commonly used record-based encoding, as Eq.~\ref{eq:record-based}: the samples HV is obtained by adding up the item HVs indexed from the item memory multiplied by the feature value. This is actually a matrix multiplication between the input feature vector and the item memory. Therefore, we can connect this to the linear layer of neural network which can be described as Eq.~\ref{eq:ffnn}, where $x$ and $y$ are the input and output of the layer, and $W$ and $B$ are the weights and biases respectively. From the perspective of HDC, the weights are essentially the item memory and the biases are set to 0. If the encoding process features more HV operations beyond multiplication, then biases can be used and multiple such layers can be cascaded.

We can also find the counterpart of activation functions in neural networks in HDC. After the encoding of input samples, HVs are often bipolarized or binarized, where the numbers larger than 0 are set to 1 and numbers smaller than 0 are set to -1 (or 0). Such bipolarization or binarization, similar to the activation functions in neural networks, provide non-linearity to HDC so that HDC models can be used to perform classification tasks, which is similar to the kernel trick~\cite{hofmann2006support} in other machine learning algorithms such as support vector machines. In this paper specifically, we apply hyperbolic tangent activation (tanh) after the input layer of the neural network which has the similar effect of bipolarization.

\begin{equation}
\vec{V} = \sum_{m = 1}^{M}{\mathbf{I}.index(f_m) \times f_m}
\label{eq:record-based}
\end{equation}

\begin{equation}
y = W^Tx + B
\label{eq:ffnn}
\end{equation}

\subsection{Associative Memory vs. Classifier Layer}
We also formulate the associative memory as the classifier layer, i.e., the output layer of a neural network. As introduced in Sec.~\ref{sec:development}, the inference process of HDC is the iterative similarity calculation between the query HV and each class HV in the associative memory. Recall the definition of cosine similarity at Eq.~\ref{eq:cosim}, we can ignore the $||\vec{V_a}||\times||\vec{V_b}||$ and simplify the calculation into the vector product. This further transforms the inference process into another matrix multiplication and can subsequently be regarded as a linear layer just like the classifier in a neural network. The activation function after the classifier layer is softmax as it resembles argmax in HDC inference which is used to locate the class with highest similarity, but is differentiable.

\section{HDC versus NN: Learning}
We have already present the architectural comparison between an HDC model. To make a further step, we also analyze in this subsection that the learning process of HDC is also similar to the back-propagation. We address that three critical characteristics of HDC learning process grant its advantages such as energy efficiency, but on the other hand limit the potential of the learning capability of high dimensional patterns for more complicated and difficult datasets.

First, for the canonical HDC model, the item memory is usually fixed. From the perspective of neural network, it means that only the classifier layer of the formulated neural networks is trainable while the input layer is ``freezed''. The item memory will not be updated during the training and retraining process. For neural network, it means the gradients are not back-propagated to the input layer and will stop at the classifier layer, which is similar to fine-tuning a retrained network model. Second, HDC training and retraining also features a ``coerce'' version of back-propagation, which is different from gradient descent: Unlike neural networks that the weights of each neuron can be updated individually and freely, for HDC, the weights representing associative memory can only be updated together, limited by the set of encoded HVs available from the training set. We recognize such training is an ``approximate back-propagation'' as the encoded HVs of each training sample contribute to rather a limited vector space. Third, HDC learning rate is much higher than that of neural networks. The canonical HDC training and retraining uses a learning rate of 1, this explains why HDC training is much faster than the neural networks as the accuracy could saturate after much less epochs.

\section{Experimental Results}
\begin{figure}
    % \centering
    \includegraphics[width = 1\columnwidth]{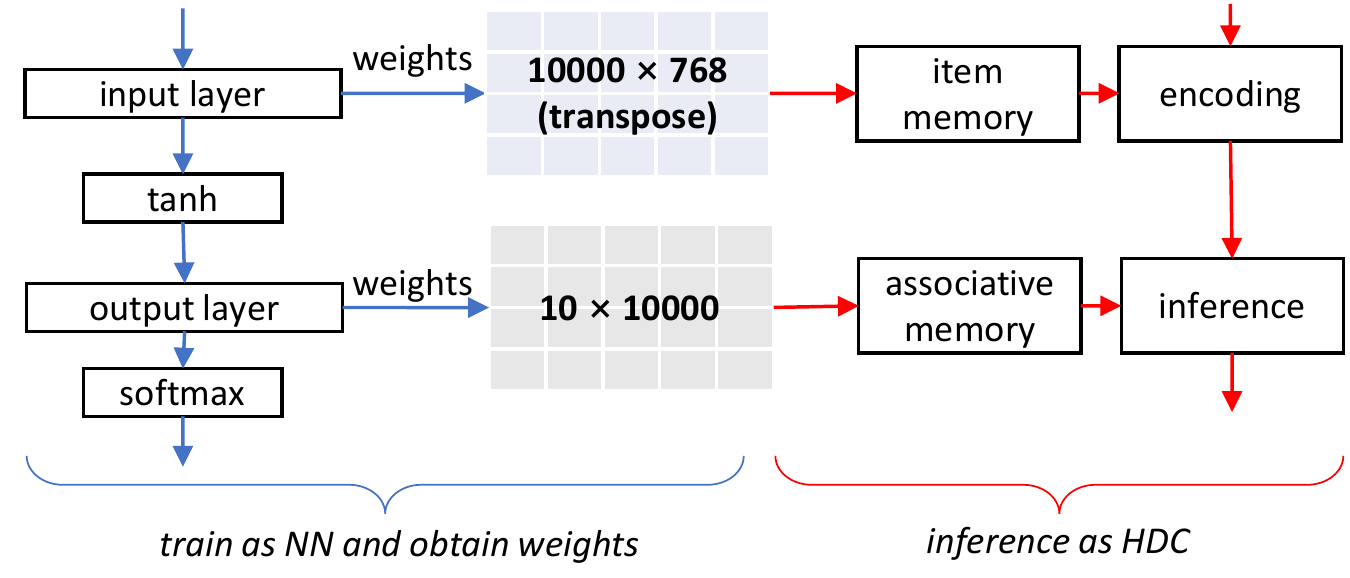}
	\caption{Derive an HDC model from a trained neural network.}
    \label{fig:derive}
\end{figure}

In this section, based on the comparison and analysis we present above, we show an experimental study that HDC model can be derived from a trained neural network which can surpass accuracy of conventional HDC models. As presented in Fig.~\ref{fig:derive}, a two-layer neural network is trained using conventional neural network training strategies. After training, the weights of each layer is fetched and then duplicated as the item and associative memory during encoding and inference.  

\subsection{Experimental Setup}

We use the MNIST dataset~\cite{lecun1998gradient} as our benchmark dataset because it is one of the datasets that HDC performs much less than state-of-the-art that most of the HDC models can only achieve less than 95\% accuracy~\cite{ma2021hdtest}. We also use CIFAR-10 dataset~\cite{krizhevsky2009learning} which SOTA HDC can only achieve around 45\% accuracy~\cite{duan2022lehdc}.

We use a typical configuration of HDC with the dimensionality of 10,000. For MNIST, we use the record-based encoding as introduced in Eq.~\ref{eq:record-based}, which translates into a neural network model with two layers as specifies by Fig.~\ref{fig:network-arch}. For CIFAR-10, as the input image has three channels (R, G and B), therefore we modify the record-based encoding as Eq.~\ref{eq:cifar-encode} so that each channel is encoded separately like MNIST dataset. After encoding all the 3 channels, the HVs representing each channel are added together each two so that different channel information can be mixed together.

We also compare the performance with two baselines:
\begin{itemize}
    \item HDC-base: This baseline is the HDC model trained using the canonical (record-based) encoding, training, and retraining process which can be found in the majority of HDC related literature.
    \item LeHDC: This baseline is implemented according to the learn-able HDC frameworks proposed in~\cite{duan2022lehdc}.
\end{itemize}

\begin{equation}
    \begin{aligned}
&        \vec{V_{RG}} = \vec{V_R} + \vec{V_G} \\
 &       \vec{V_{RB}} = \vec{V_R} + \vec{V_B} \\
  &      \vec{V_{GB}} = \vec{V_G} + \vec{V_B} \\
   &     \vec{V} = \vec{V_{RG}} + \vec{V_{RB}} + \vec{V_{GB}}
    \end{aligned}
    \label{eq:cifar-encode}
\end{equation}

\subsection{Training of Neural Network}
We implement the neural network using PyTorch framework, the HDC network architecture can be concisely defined as a sequential model of PyTorch with just a few lines along with the required activation functions. For example, the code for defining HDC model for MNIST dataset is shown in Listing~\ref{list:mnist} as an example. We train the neural network with Adam optimizer and use learning rate 0.001 and the training is terminated if inference accuracy does not substantially increase after consecutive epochs.

\begin{lstlisting}[language=Python, caption={Defining HDC Model for MNIST using PyTorch}, label={list:mnist}]
    def hdc(d_feature, d_HV, n_classes):
      return nn.Sequential(
        nn.Flatten(),
        nn.Linear(d_feature, d_HV, bias = False),
        nn.Tanh(),
        nn.Linear(d_HV, n_classes, bias = False),
        nn.Softmax(dim = 1))
\end{lstlisting}

\subsection{Derive an HDC from NN}
We derive an HDC model from a trained network by extracting the weights of the input and classifier layers. For the input layer, the size of the weight matrix is $10000\times768$, thus each column can be considered as an item HV and the entire weight matrix, with a transpose, is thus the item memory. For the classifier layer as the weight matrix size is $10\times10000$, the row instead can be considered as the class HV thus the weight matrix can be considered as the associative memory.

We present an accuracy comparison of the two baseline models and the HDC model transformed from the trained neural network in Table~\ref{tab:accuracy}. We can observe that for the MNIST dataset, the HDC model derived from neural network can achieve about 96.7\% accuracy which surpasses the baseline by nearly 6\% and the learning based LeHDC by about 2\%.

For the even more challenging CIFAR-10 dataset, the canonical \textbf{HDC-base} model can only achieve unacceptable accuracy of 30\%. For the LeHDC baseline, the accuracy can increase to 46\%. Although for this paper with HDC model transformed from neural network, the accuracy is still only 51\%, there is already a 20\% increase from the baseline.

As to computation cost, the HDC model transformed from the neural network performs the same amount of computations as the HDC-Base during inference, since they share the same architecture. The overhead is on the training process since a neural network needs to be trained to obtain the parameters such as weights to be transformed into the HDC memory elements. Additionally, since the neural network architecture trained for transform is compact (as shown in Listing~\ref{list:mnist} ), the training is also fast. With our experimental setup, it takes around 4 and 7 minutes to achieve the reported accuracy even when trained afresh.

\begin{table}[htbp]
  \centering
  \caption{Accuracy Comparison with Two Baseline Models}
    \begin{tabular}{cccc}
    \toprule
          & HDC-Base & LeHDC~\cite{duan2022lehdc} & This Paper \\
    \midrule
    MNIST & $90.93^{\pm 0.52}$  & $94.74^{\pm 0.18}$ & $96.71^{\pm 0.37}$ \\
    CIFAR-10 & $30.28^{\pm 1.90}$  & $46.10^{\pm 0.20}$  & $51.08^{\pm 0.79}$ \\
    \bottomrule
    \end{tabular}%
  \label{tab:accuracy}%
\end{table}%

\subsection{Discussion}
HDC resembles the architecture of a extremely compact neural network with just 2 layers. Both the encoding of HV and the inference can be accomplished by a typical fully connected layer. The item memory and the associative memory can be transformed from the weights of a trained neural network in such an architecture. The training of HDC models is also similar to the training of a neural network. Instead of using back-propagation with gradient descent, HDC use HV addition and subtraction to guide model to converge at the direction of higher accuracy which is easier to implement. We recognize this as a compromise to sacrifice some of the canonical neural network training schemes for more energy efficiency and acceleration. 

The learning capability of HDC is capped by the corresponding neural network architecture. From the experimental results, although a two-layer neural network can achieve acceptable accuracy on the MNIST, when the dataset becomes much complicated and challenging like the CIFAR-10, such shallow network structure cannot achieve an acceptable accuracy. This explains the reason that the current HDC models mostly focus on simple applications or datasets and cannot achieve comparable performance on more demanding learning tasks. 

One major challenge of HDC is that the encoding process is not application-agnostic. System designers are required to spend manual effort to design, develop, and evaluate the encoding process, which can often lead to sub-par results as there may exist undiscovered encoding methods with better performance and also prohibits the scalability and flexibility. Neural network has the advantage over HDC as the weights of neural network is individually trainable, however, in HDC only the associative memory is trainable and the elements inside each class HV is not individually trainable due to the holographic representation of HVs.

\section{Conclusion}
The tremendously increasing popularity of Hyperdimensional Computing (HDC) has attracted researchers from various domains to invest their effort in this topic. The major advantages of HDC are better energy efficiency, smaller model size, and acceleration on heterogeneous platforms. In this paper, we provide a new perspective of getting insights of this emerging algorithm from the angle of neural networks. Specifically, we present a comparative analysis and experiment to illustrate the similarity between HDC and neural network on architecture and learning process, and show that an HDC model can be derived from a two-layer neural network. During experiments, we illustrate that by training the neural network upfront and then derive an HDC model based on the trained network, we can achieve up to 21\% accuracy improvement from baseline HDC models and up to 5\% improvement from SOTA learning based HDC models. 
\bibliography{HDCNN}
\end{document}